\documentclass[letterpaper]{article} % DO NOT CHANGE THIS
\usepackage{aaai25}  % DO NOT CHANGE THIS
\usepackage{times}  % DO NOT CHANGE THIS
\usepackage{helvet}  % DO NOT CHANGE THIS
\usepackage{courier}  % DO NOT CHANGE THIS
\usepackage[hyphens]{url}  % DO NOT CHANGE THIS
\usepackage{graphicx} % DO NOT CHANGE THIS
\usepackage{natbib}  % DO NOT CHANGE THIS AND DO NOT ADD ANY OPTIONS TO IT
\usepackage{caption} % DO NOT CHANGE THIS AND DO NOT ADD ANY OPTIONS TO IT
\usepackage{algorithm}
\usepackage{algorithmic}
\usepackage{bm}
\usepackage{subcaption}
\usepackage{mathtools}
\usepackage{nicematrix}
\usepackage{subcaption}
\usepackage{multirow}
\usepackage{color, colortbl}
\usepackage{amsmath}
\usepackage{amssymb}
\usepackage{pifont}
\newcommand{\yes}{\text{\ding{51}}}
\newcommand{\no}{\text{\ding{55}}}
\usepackage{booktabs}
\definecolor{defaultcolor}{gray}{.9}
\newcommand{\default}[1]{\cellcolor{defaultcolor}{#1}}
\usepackage[capitalize]{cleveref}
\crefname{section}{Sec.}{Secs.}
\crefname{table}{Tab.}{Tables}
\crefname{figure}{Fig.}{Figs.}

\usepackage{newfloat}
\usepackage{listings}
\DeclareCaptionStyle{ruled}{labelfont=normalfont,labelsep=colon,strut=off} % DO NOT CHANGE THIS
\floatstyle{ruled}
\newfloat{listing}{tb}{lst}{}
\floatname{listing}{Listing}
\nocopyright % -- Your paper will not be published if you use this command
\title{MambaVT: Spatio-Temporal Contextual Modeling for robust RGB-T Tracking}
\author{
    Simiao Lai\textsuperscript{\rm 1}, Chang Liu\textsuperscript{\rm 1}, Jiawen Zhu\textsuperscript{\rm 1}, Ben Kang\textsuperscript{\rm 1}, Yang Liu\textsuperscript{\rm 1}, Dong Wang\textsuperscript{\rm 1}, Huchuan Lu\textsuperscript{\rm 1}
}
\affiliations{
    \textsuperscript{\rm 1}Dalian University of Technology\\
    laisimiao@mail.dlut.edu.cn
}

\usepackage{bibentry}
\begin{document}

\maketitle
%%%%%%%%% ABSTRACT
\begin{abstract}
Existing RGB-T tracking algorithms have made remarkable progress by leveraging the global interaction capability and extensive pre-trained models of the Transformer architecture. 
Nonetheless, these methods mainly adopt image-pair appearance matching and face challenges of the intrinsic high quadratic complexity of the attention mechanism, resulting in constrained exploitation of temporal information.
Inspired by the recently emerged State Space Model Mamba, renowned for its impressive long sequence modeling capabilities and linear computational complexity, this work innovatively proposes a pure Mamba-based framework (\textbf{MambaVT}) to fully exploit spatio-temporal contextual modeling for robust \textbf{v}isible-\textbf{t}hermal tracking.
Specifically, we devise the long-range cross-frame integration component to globally adapt to target appearance variations, and introduce short-term historical trajectory prompts to predict the subsequent target states based on local temporal location clues. 
Extensive experiments show the significant potential of vision Mamba for RGB-T tracking, with MambaVT achieving state-of-the-art performance on four mainstream benchmarks while requiring lower computational costs.
We aim for this work to serve as a simple yet strong baseline, stimulating future research in this field. 
The code and pre-trained models will be made available.
\end{abstract}

%M$^2$-Track
%%%%%%%%% INTRODUCTION
\section{Introduction}\label{sec:intro}
%%%%%%%%%%%%%%%%% rgbt tracking %%%%%%%%%%
RGB-T tracking aims to estimate the target's position in subsequent video frames using RGB and thermal infrared images, given the initial target state in the first frame. Compared to RGB-based tracking, thermal infrared images can perceive targets with different thermal radiation and provide complementary information that enhances the tracker's robustness in all-weather conditions, particularly in extreme-illumination and low-texture environments. This has attracted researchers to make significant progress in multi-modal feature extraction\cite{cat++,CMPP} and interactive fusion\cite{DAFNet,APFNet,tbsi}, advancing the field of RGB-T tracking.
%%%%%%%%%%%%%%%%%%%%%%%%%
\begin{figure}[!t]
\centering
\includegraphics[width=0.97\linewidth]{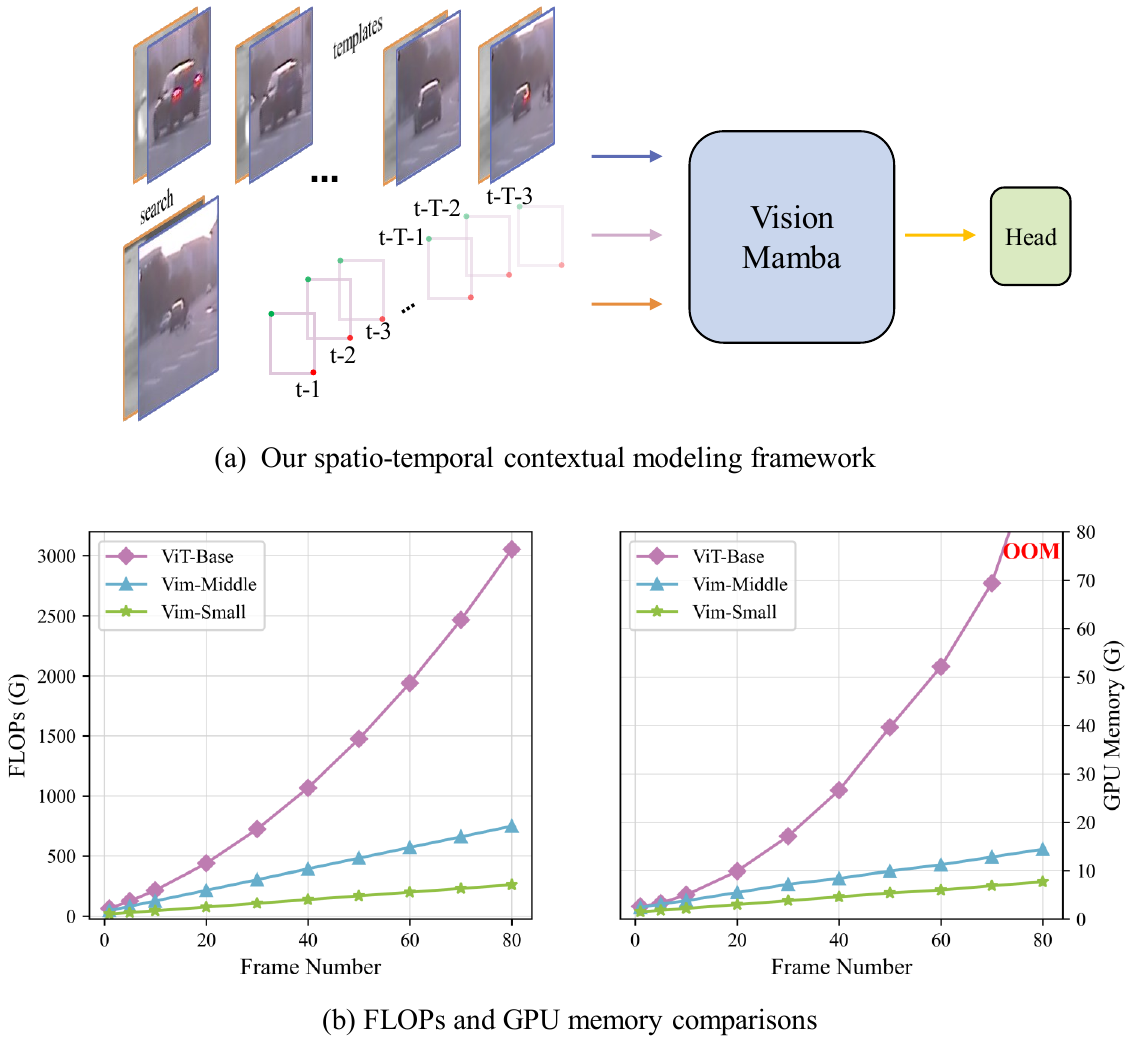}
	\caption{Framework and efficiency comparisons of our proposed methods. (a) Unlike existing RGB-T tracking methods that usually take a single image pair as input and overlook temporal information, our framework uses vision Mamba to fully exploit the spatio-temporal contexts from the perspective of long-range appearance modeling and short-term motion modeling. (b) The FLOPs and GPU memory usage of transformer-based methods grow quadratically as the number of frames increases and become unbearable. In contrast, our Mamba-based method scales linearly, making it efficient. Note that batch size is 1.}
	\label{fig:firstpage}
\end{figure}

%%%%%%%%%%%%% transformer works and disadvantages  %%%%%%%%%%%%
Transformer\cite{transformer} is emerging as the dominant architecture in RGB-T tracking, owing to their global receptive field and capacity to process variable-length input sequences.
Many endeavors\cite{ProFormer,MMMPT,CSTNet} have employed the transformer encoder and attention mechanism for cross-modal feature fusion. 
Furthermore, USTrack\cite{USTrack} embeds the template and search images from two modalities into patch tokens, and simply concatenates them for input into vision transformer\cite{dosovitskiy2020vit}, which enables joint modality features extraction and mutual interaction within a single-stage pipeline.
Moreover, the ViTs community offers a rich and powerful set of pre-trained models, which facilitates many works\cite{protrack,vipt} in utilizing visual prompt learning paradigm to incorporate auxiliary modal information into the foundation model trackers, thereby further enhancing tracking performance.
Despite the considerable success of transformer-based methods, exploration of temporal information in the RGB-T tracking field remains limited.
Most methods still adopt the image-level manner to locate the objects by matching the template appearance feature in the search region.
We deduce this is primarily due to the conflict between the image-pair matching tracking scheme, which lacks long sequence context correlation modeling, and the inherent quadratic complexity of transformer attention mechanism.
As illustrated in \cref{fig:firstpage}(c), with the increase in the number of frames (same as sequence length), both the computational operations and GPU memory consumption of ViT exhibit quadratic growth. This poses a significant computational challenge for leveraging long-range temporal information.
%
%%%%%%%%%%%%%%%%%%%%%%%%%

%%%%%%%%%%%%% our methods and introduction %%%%%%%%%%
To alleviate this contradiction, we turn our attention to Mamba\cite{gu2023mamba}. Recently, the Selective Structured State Space Model has demonstrated significant potential in modeling long-range dependencies with linear complexity.
Pioneering works like Vim\cite{vim} and VMamba\cite{liu2024vmamba} first validate its effectiveness in classification tasks and downstream dense prediction tasks.
Afterward, Mamba has emerged with remarkable superiority across a wide range of visual tasks, especially those requiring long sequence modeling, such as video understanding\cite{li2024videomamba, yang2024vivim} and high-resolution medical image processing\cite{xing2024segmamba,liu2024swinumamba,vmunet}.
These successful applications have encouraged us to adapt Mamba to RGB-T multi-modal video tracking task, particularly leveraging its linear complexity long-sequence modeling capability to comprehensively model spatio-temporal context within reasonable computational constraints, thereby enhancing the robustness and overall performance.
Hence, inheriting the one-stream architecture, we introduce a compact Mamba-based RGB-T tracking framework, consisting of a long-range cross-frame integration component and short-term historical trajectory prompts.
The former integrates multi-template information across the entire video, caching past targets during tracking to globally handle object deformation and appearance variations.
The latter learns to locally estimate the current target position using motion cues based on the historical states in the short neighborhood of the current frame.
To this end, relying on Mamba's hardware-aware algorithm, our MambaVT can still operate in real-time despite the long-sequence input, and the small variant achieves an astonishing 57.9\% AUC performance on LasHeR\cite{li2021lasher} with relatively low computational cost.

The contributions of this work are summarized as follows:
\begin{itemize}
\item To the best of our knowledge, we are the first to leverage the linear long sequence modeling merits of Mamba to explore spatio-temporal contextual modeling for RGB-T tracking, providing a simple and effective pure Mamba-based baseline.

\item Within a compact framework, we propose the long-range cross-frame integration component and short-term historical trajectory prompts to model contexts both globally and locally, achieving a more comprehensive exploitation of spatio-temporal information.

\item Benefiting from the Mamba architecture, our MambaVT shows superiority on multiple RGB-T tracking datasets against the state-of-the-arts, even with fewer computational costs and lower memory requirements.

\end{itemize}
%%%%%%%%% RELATED
\section{Related Works}\label{sec:related}

\subsection{RGB-T Tracking}
RGB-T trackers are designed to utilize thermal infrared modality information to complement RGB images, thereby enhancing the algorithm's robustness in challenging scenarios such as low-light night-time and backlit environments. 
Consequently, many earlier works \cite{mfDiMP, cat++, APFNet, hmft} focused on developing sophisticated and elaborate fusion modules or strategies to extract and integrate features from different modalities. 
Recently, the global perception capability of vision transformer\cite{dosovitskiy2020vit} and their abundant pre-trained representation ecosystem have driven researchers\cite{vipt,UnTrack,sdstrack,onetracker} to use visual prompt learning to inject auxiliary modality information into foundational models, further improving RGB-T trackers' performance.
However, the inherent quadratic computational complexity of the self-attention mechanism in transformer has hindered the development of long-range temporal modeling in transformer-based RGB-T trackers. 
Although TATrack\cite{tatrack} utilizes the initial branch and the online branch to enable cross-modal interaction over longer time scales, it only models the appearance variation of a single online template, lacking the capability for long-range cross-frame appearance modeling and motion modeling.
In this work, inspired by Mamba's efficient long-sequence modeling property, we introduce Mamba into the RGB-T tracking domain for the first time, modeling the spatio-temporal context dependencies to enhance tracking robustness. We hope this work serves as a simple and effective baseline to stimulate future research.

\begin{figure*}[!t]
\centering
\includegraphics[width=0.99\textwidth]{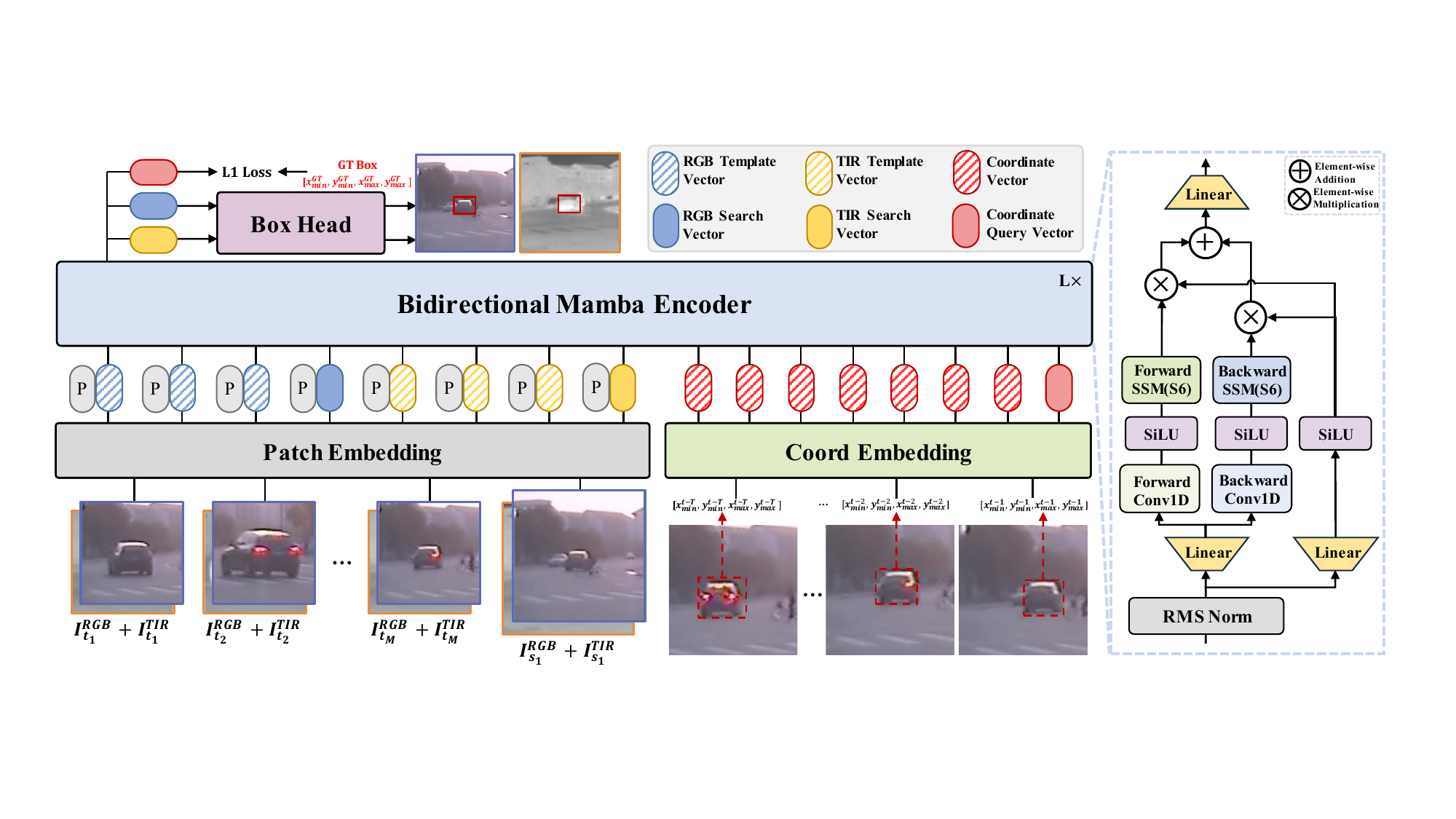}
	\caption{Overall framework of our MambaVT. The video-level templates set and search region training samples are fed into patch embedding layer, and historical trajectory prompts are fed into coordinate embedding layer. The embedded vectors are all sent into bidirectional Mamba encoder for unified contextual modeling. Ultimately, the search region vectors are used for predicting object state and coordinate query vector is used for auxiliary supervision with ground-truth bounding box.}
	\label{fig:frame}
\end{figure*}

\subsection{Vision Mamba Models}
Compared to previous structured State Space Models\cite{S4}, Mamba leverages an input-dependent selection mechanism and a hardware-aware parallel algorithm\cite{gu2023mamba}, allowing it to model long-range dependency linearly with sequence length.
Initially, it achieves outstanding performance in continuous long-sequence data analysis, such as natural language processing. 
Recently, Mamba's linear complexity in long-range modeling has proven effective and superior in many visual tasks. Vim\cite{vim} and VMamba\cite{liu2024vmamba} are pioneers in applying Mamba to the classification task.
This success has led to its adoption in subsequent high-resolution image tasks, including notable works in medical image segmentation like VM-UNet\cite{vmunet}, SegMamba\cite{xing2024segmamba}, and Swin-UMamba\cite{liu2024swinumamba}, as well as in remote sensing image with RSMamba\cite{chen2024rsmamba} and ChangeMamba\cite{chen2024changemamba}. In the video domain, VideoMamba\cite{li2024videomamba} has shown significant potential in understanding both short-term and long-term video contents. Vivim\cite{yang2024vivim} presents a generic video vision Mamba framework that utilizes hierarchical temporal Mamba blocks to compress cross-frame temporal dependencies. MambaTrack\cite{MambaTrack} explores using Mamba as a learning-based motion model for MOT. In contrast, our work is the first to investigate and demonstrate Mamba's long-range contextual modeling capability in RGB-T multi-modal single object tracking.

%%%%%%%%% METHOD

\section{Methodology}\label{sec:methodology}
In this section, we detail the proposed MambaVT. First, we introduce the preliminaries of state space models and the Mamba.
Then, we leverage the long-sequence modeling capacity of Mamba to construct the spatio-temporal correlations within a compact framework from two viewpoints: a global view of long-range cross-frame integration and a local view of short-term historical trajectory prompts.
Last, we present the strategies to update templates online.
The overall architecture of MambaVT is shown in \cref{fig:frame}.

%%%%%%%%%%%%%%%%%%%%%%
%%%%%%%%%%%%%%%%%%%%%%
\subsection{Preliminaries}
Since \cite{S4} propose the Structured State Space (S4) model that solves the SSM's problems of the prohibitive computation and memory requirements, the SSMs have emerged as a promising class of architectures for deep sequence modeling.
The SSM-based models are derived from a continuous system that maps the 1-dimensional sequence $x(t) \in \mathbb{R}^{L} \mapsto y(t) \in \mathbb{R}^{L}$ through a hidden state $h(t) \in \mathbb{R}^{N}$. They define the sequence-to-sequence transformation with evolution parameter $\mathbf{A} \in \mathbb{R}^{N \times N}$ and projection parameters $\mathbf{B} \in \mathbb{R}^{N \times 1}$ and $\mathbf{C} \in \mathbb{R}^{1 \times N}$ as follows:
\begin{equation}
\begin{aligned}
h^{\prime}(t) & =\mathbf{A} h(t)+\mathbf{B} x(t) \\
y(t) & =\mathbf{C} h(t)
\end{aligned}
\end{equation}
The Modern SSMs, \textit{i.e.}, S4 and Mamba are the discrete forms of this continuous state. The process of discretization is typically accomplished using a rule called zero-order hold (ZOH):
\begin{equation}
    \begin{aligned}
&\overline{\mathbf{A}}=\exp{(\Delta\mathbf{A})},\\
&\overline{\mathbf{B}}=(\Delta\mathbf{A})^{-1}(\exp{(\Delta\mathbf{A})}-\mathbf{I})\cdot\Delta\mathbf{B}. \\
&h_{t} =\overline{\mathbf{A}}h_{t-1}+\overline{\mathbf{B}}x_{t}, \\
&y_{t} =\mathbf{C}h_{t}. 
\end{aligned}
\label{eq:S6}
\end{equation}
where $\Delta$ is a timescale parameter, $h_{t}$, and $h_{t-1}$ symbolize the discrete hidden states at various time steps. $\overline{\mathbf{A}}$ and $\overline{\mathbf{B}}$ are the discrete counterparts of parameters $\mathbf{A}$ and $\mathbf{B}$. Moreover, Mamba stands out from previous time-invariant structured SSMs by incorporating a selective scan mechanism (S6) that endows the model with time-varying and context-aware capabilities. This is achieved by parameterizing the SSM parameters $\mathbf{B} \in \mathbb{R}^{B \times L \times N}$, $\mathbf{C} \in \mathbb{R}^{B \times L \times N}$ and $\Delta \in \mathbb{R}^{B \times L \times D}$ using linear projection based on the input $x \in \mathbb{R}^{B \times L \times D}$.

%%%%%%%%%%%%%%%%%%%%%%
%%%%%%%%%%%%%%%%%%%%%%
\subsection{Long-range Cross-frame Integration}
To adapt to the dynamic changes of the target appearance during tracking, we propose the integration component of long-range cross-frame spatial and temporal information to provide a more robust and enriched appearance reference. 
In this work, we inherit the merits of transformer-based one-stream network for joint and efficient feature extraction and information interaction. Thus, we adopt a plain and compact one-stream structure by simply stacking vision Mamba blocks. 
Let $\{\bm{I}_{t_i}^{RGB},\bm{I}_{t_i}^{TIR} \in \mathbb{R}^{H_{z} \times W_{z} \times 3} \mid i \in \{1,2,\ldots,\mathcal{M}\} \}$ denote the RGB and TIR template frame sets and $\bm{I}_{s_1}^{RGB},\bm{I}_{s_1}^{TIR}$ denote the RGB and TIR search region frames respectively, where $\mathcal{M}$ is the total template number in the set. These frames are sampled from the same video.
As illustrated in \cref{fig:frame}, these video-level frames are firstly projected into spatiotemporal patches through the path embedding layer that consists of 2-dimensional convolution with the same kernel size and strides of $P \times P$.
Similar to training of ViT, we add learnable positional encodings $\bm{P}_{t} \in \mathbb{R}^{L_{z} \times D},\bm{P}_{s} \in \mathbb{R}^{L_{x} \times D}$ to the embedded template vectors $\bm{V}_{t}^{RGB},\bm{V}_{t}^{TIR} \in \mathbb{R}^{(ML_{z}) \times D}$ and search region vectors $\bm{V}_{s}^{RGB},\bm{V}_{s}^{TIR} \in \mathbb{R}^{L_{x} \times D}$ to provide the positional prior, where $L_{z},L_{x}$ denote the length of template vectors and search region vectors.
All vectors are then concatenated along the length dimension and processed together through stacked $L$ layers of bidirectional vision Mamba encoders $E^{l}$:
\begin{equation}
\begin{aligned}
\bm{V}^{0}_{ts} & = [\bm{V}_{t}^{RGB};\bm{V}_{s}^{RGB};\bm{V}_{t}^{TIR};\bm{V}_{s}^{TIR}] \\
\bm{V}^{l+1}_{ts} & = E^{l+1}(\bm{V}^{l}_{ts}),\quad l=0,1,\ldots,L-1
\end{aligned}
\label{eq:pe}
\end{equation}
The detailed structure of bidirectional vision Mamba encoders $E^{l}$ is shown on the right side of \cref{fig:frame}.
The input vector sequence $\bm{V}^{l}_{ts}$ first undergoes normalization through a normalization layer, and then is individually processed by two linear layers to increase the dimension $D$ to the inner dimension $D^{inner}$. The intermediate features obtained from this process are denoted as $\bm{Z}_{ts},\bm{X}_{ts} \in \mathbb{R}^{(2ML_{z}+2L_{x}) \times D^{inner}}$:
\begin{equation}
\begin{aligned}
\bm{Z}_{ts} & = Linear^{up}_{z}(RN(\bm{V}^{l}_{ts})) \\
\bm{X}_{ts} & = Linear^{up}_{x}(RN(\bm{V}^{l}_{ts}))
\end{aligned}
\end{equation}
where RN and $Linear^{up}$ represent RMS normalization\cite{zhang2019rmsnorm} and dimension-increasing linear layer.
\begin{equation}
\begin{aligned}
\bm{Y}_{o} & = S6_{o}(SiLU(Conv1d_{o}(\bm{X}_{ts}))) \\
\bm{Y}^{\prime}_{o} & = \bm{Y}_{o} \odot SiLU(\bm{Z}_{ts}) \\
\bm{V}^{l+1}_{ts} & = Linear^{down}(\bm{Y}^{\prime}_{\mathrm{forward}}+\bm{Y}^{\prime}_{\mathrm{backward}})
\end{aligned}
\end{equation}
where subscript $o$ denotes the two scan orientations, \textit{i.e.}, forward and backward scan. Bidirectional scanning facilitates mutual interactions among all elements within the sequence, thereby establishing a global and unrestricted receptive field. $Linear^{down}$ is the linear layer for reducing the inner dimension $D^{inner}$ back to the original dimension $D$. The information flow of $S6$ is depicted in \cref{eq:S6}.

Afterward, the search region vectors of RGB and TIR modalities from the output of the last encoder $E^{L}$ are added element-wisely and sent into the box head to predict the target states. Here, we implement the center-based head\cite{ostrack} that consists of stacked $Conv\text{-}BN\text{-}ReLU$ layers. 
To sum up, training with video-level input across frames enables the model to leverage long-range spatio-temporal contexts, integrating and memorizing diverse reference appearances to identify the target's location in the current frame.
Accordingly, we concatenate both the patch embedding vectors and coordinate embedding vectors and feed them into the bidirectional Mamba encoder for unified contextual modeling. 

%%%%%%%%%%%%%%%%%%%%%%
%%%%%%%%%%%%%%%%%%%%%%
\subsection{Short-term Historical Trajectory Prompts}
Although multi-frame input provides the model with a global perspective spanning the video, the model still lacks robustness when handling moving targets that are affected by similar backgrounds or occluded. Therefore, to complement this shortcoming, we incorporate a mechanism that uses short-term historical trajectory prompts to improve the model's tracking performance in complex environments.
As illustrated in \cref{fig:frame}, we first need to convert continuous coordinates into discrete vector sequences using a shared vocabulary.
Specifically, assuming the index of the current search region frame is $t$, we sample the tracking results of the previous consecutive $\mathcal{T}$ frames (a local neighborhood) as trajectory prompts:
\begin{equation}
\{[x_{min}^{t-j},y_{min}^{t-j},x_{max}^{t-j},y_{max}^{t-j}] \mid j \in \{1,2,\ldots,\mathcal{T}\} \}
\end{equation}
Then, to be consistent with subsequent loss calculation and supervision, every continuous coordinate of different frames should be mapped into the same coordinate system, \textit{i.e.}, the current search region coordinate system.
Next, every coordinate is uniformly discretized into an integer between $[1, nbins]$. Each integer acts as an index to retrieve a coordinate embedding vector $\bm{V}^{0}_{coord} \in \mathbb{R}^{1 \times D}$ from the vocabulary.
Following ARTrack\cite{artrack}, we also adopt a coefficient of dilatation $\alpha$ to expand the actual representation range of the vocabulary and include more effective preceding trajectory sequences. 
Additionally, to encourage the model to utilize trajectory information during training, we add a coordinate query vector $\bm{V}^{0}_{query} \in \mathbb{R}^{4 \times D}$ that is used to compute the L1 loss against the ground truth boxes for auxiliary supervision.
Accordingly, we concatenate both the patch embedding vectors and coordinate embedding vectors and feed them into the bidirectional Mamba encoder for unified contextual modeling. 
The \cref{eq:pe} is now reformulated as follows:
\begin{equation}
[\bm{V}^{l+1}_{ts};\bm{V}^{l+1}_{coord};\bm{V}^{l+1}_{query}] = E^{l+1}([\bm{V}^{l}_{ts};\bm{V}^{l}_{coord};\bm{V}^{l}_{query}])
\end{equation}
To this end, we add a local motion-based inference view to supplement the global appearance-based inference for robust tracking. 
And Mamba's specialty of scaling linearly with sequence length promises satisfactory computational cost even with multi-modal and multi-source input flows.

%%%%%%%%%%%%%%%%%%%%%%
%%%%%%%%%%%%%%%%%%%%%%
\subsection{Online Template Memory Selection}
Rather than retraining a light classification branch to estimate the probability of foreground quality\cite{yan2021stark,cui2022mixformer}, or using complicated heuristic strategies to dynamically adjust thresholds\cite{tatrackrgb}, we adopt a simple yet empirically proven effective strategy for updating templates. As in ODTrack\cite{zheng2024odtrack}, we use the following formula to determine the index of the template being selected:
{\small
\begin{equation}
\begin{cases} 
\{0\} \cup \left\{ \left( i \cdot K + \left\lfloor \frac{K}{2} \right\rfloor \right) \mid i \in \{0,1,\ldots,\mathcal{M}-1\} \right\}, & \text{if } \mathcal{M} > 1 \\
\{0\}, & \text{if }\mathcal{M}=1
\end{cases}
\label{eq:update}
\end{equation}
}
Where $K=\left\lfloor\frac{C_{i}}{\mathcal{M}}\right\rfloor$ denotes the average memory duration of each template. $C_{i}$ represents the index of the current frame.

This formula not only considers the initial frame template, which contains the most accurate target appearance but also includes targets from later periods close to the current frame as dynamic templates. It spans the entire tracked video, achieving long-range cross-frame information integration during testing, consistent with the training process. Our experiments reveal that although this approach might allow lower-quality templates into the template set, the results remain promising. We deduce that by utilizing multiple historical templates, even if a few inferior templates are incorporated, the remaining templates can still maintain an accurate overall target judgment. 

%%%%%%%%% EXPERIMENTS
\section{Experiments}\label{sec:experiments}
\subsection{Implementation Details}

{\flushleft\textbf{Model Variants.}}\quad
We trained two variants of MambaVT with different configurations as follows:
\begin{itemize}
\item \textbf{$\text{MambaVT-S}_\text{256}$.} Backbone: Vim-Small; Template size: [128\begin{math}\times\end{math}128]; Search region size: [256\begin{math}\times\end{math}256];
\item \textbf{$\text{MambaVT-M}_\text{256}$.} Backbone: Vim-Middle; Template size: [128\begin{math}\times\end{math}128]; Search region size: [256\begin{math}\times\end{math}256];
\end{itemize}
% TODO: 这里缺少一些深度和维度的信息，看吧，如果空间不够就写到补充材料里面

{\flushleft\textbf{Training.}} We implement our MambaVT based on  PyTorch2.1 and train it on 2 NVIDIA A100 SXM 80GB GPUs, with a batch size of 32 on each card. To align with the existing transformer-based methods, we first take the videomamba\cite{li2024videomamba} as initial parameters and pre-train the MambaVT like OSTrack\cite{ostrack} on the widely-used tracking data, such as COCO\cite{coco}, LaSOT\cite{lasot}, GOT10k\cite{got10k}, and TrackingNet\cite{trackingnet}. Then, the whole fine-tuning on the LasHeR training set is divided into two stages: during the first stage, we input multiple template images and one search region to conduct long-range cross-frame appearance modeling. This stage involves training for 20 epochs, each comprising $6\times 10^{4}$ samples, with an initial learning rate of $8\times 10^{-4}$. The backbone network is further multiplied by a factor of 0.05. In the second stage, leveraging the results inferred by the model trained in the first stage, we incorporate the historical coordinates of the previous 7 frames of the current frame as motion prompts, enabling the model to learn short-term trajectory inference. This stage consists of 10 epochs, with an initial learning rate of $8\times 10^{-5}$, while other settings remain consistent with the first stage.

{\flushleft\textbf{Inference.}} In the inference phase, we commence by initializing a template set and a historical position queue, anchored on the initial ground truth. Leveraging all accumulated historical templates along with trajectory motion clues, the model infers the results for the current frame and decides whether to update the template based on \cref{eq:update}. However, the state of each frame is enqueued into the position queue in a first-in-first-out manner. When the queue is full, the state of the head element will be dequeued. Refer to the supplementary materials for more training and inference details.

\subsection{Main Results}
We evaluate and compare our proposed $\text{MambaVT-S}_\text{256}$ and $\text{MambaVT-M}_\text{256}$ with previous state-of-the-art RGB-T tracking methods on several popular benchmarks, including
GTOT\cite{gtot}, RGBT210\cite{rgbt210}, RGBT234\cite{rgbt234} and LasHeR\cite{li2021lasher}. 

\textbf{GTOT.} GTOT\cite{gtot} is an RGB-T dataset containing 50 video sequences under different scenarios and conditions, with a total of about 15K frame pairs. precision rate (PR) and success rate (SR) are adopted as the evaluation metrics. As shown in \cref{tab:comparison}, our MambaVT-M achieves 95.2\% PR and 78.6\% SR, 0.1\% higher on SR than the previous best tracker GMMT, and we surpass it by 1.6\% on PR, which demonstrates the superior robustness of continuous tracking without drifting, benefiting from the spatio-temporal contextual modeling.

\textbf{RGBT210.} Compared to GTOT, RGBT210\cite{rgbt210} provides more videos capturing dynamic scenes and more evaluation sequences. As shown in \cref{tab:comparison}, both variants of our method exceed 63\% on SR and 88\% on PR, outperforming previous state-of-the-art trackers. Moreover, the smaller variant behaves well at scale regression, while the larger model shows enhanced robustness in localization. These results demonstrate the effectiveness of our approach.

\textbf{RGBT234.} RGBT234\cite{rgbt234} is a large-scale RGB-T tracking benchmark, which contains 234 visible and thermal infrared videos with about 234K frames5 in total. Maximum Success Rate (MSR) and Maximum Precision Rate (MPR) are adopted for evaluation due to small alignment errors. As shown in \cref{tab:comparison}, our MambaVT-S achieves 88.9\% MPR and 65.8\% MSR, surpassing the MPR and MSR scores of the transformer-based temporal algorithm TATrack by 1.7\% and 1.4\%, respectively. Our middle version improves upon the small version by 1.7 percentage points on MSR and 1.8 percentage points on MPR, achieving top performance.

\textbf{LasHeR.} LasHeR\cite{li2021lasher} is a more modern and challenging benchmark for short-term RGB-T tracking, with up to 19 attributes to characterize different scenarios. We report the success rate, precision rate, and normalized precision rate on 245 testing videos. Surprisingly, our small version even outperforms the larger one and surpasses the previous SOTA algorithm GMMT by a substantial margin, reaching the top SR, NPR, and PR of 57.9\%, 69.5\%, and 73.0\%, respectively. For more comparative results, please refer to the supplementary materials.

\textbf{Speed and parameters.} As demonstrated in \cref{tab:comparison}, MambaVT-S can run in real-time at more than 45 fps on an RTX 3090 GPU. 
Besides, the parameters of MambaVT-S are 3$\times$ less than the parameters of SeqTrackV2. 
It is worth noting that, compared to other methods that use a pair of template and search images as input, the inference speed of our method is measured with seven template images as input. 
Despite the sequence-level inputs, our MambaVT-M maintains real-time speed at approximately 30 fps. 
Moreover, it has a smaller parameter count compared to most competitors developed on the ViT Base backbone, achieving a good balance between performance and efficiency.

%%%%%%%%%%%%%%%%%%%%%%%%%%%%%%%%%%%%%%%%%%%
\begin{table*}[!t]
\renewcommand{\arraystretch}{1}
\setlength{\tabcolsep}{6pt}
%\vspace{-6pt} @{\hspace{-0.pt}}  @{\hspace{-0.7pt}}
\centering
{
\resizebox{\textwidth}{!}{
\begin{NiceTabular}{r|cc|cc|cc|ccc|c|c}
% \begin{tabular}{@{}r|cc|ccccccc@{}}
\toprule
\multirow{2}{*}{Method} &\multicolumn{2}{c}{GTOT} &\multicolumn{2}{c|}{RGBT210}
&\multicolumn{2}{c|}{RGBT234} &\multicolumn{3}{c}{LasHeR} & \multirow{2}{*}{Params$\downarrow$} & \multirow{2}{*}{FPS$\uparrow$} \\
 &PR$\uparrow$ &SR$\uparrow$ &PR$\uparrow$ &SR$\uparrow$ &MPR$\uparrow$ &MSR$\uparrow$ &PR$\uparrow$ &NPR$\uparrow$ &SR$\uparrow$ & & \\
\midrule
mfDiMP~\cite{mfDiMP} &83.6&69.7&78.6&55.5&-&-&-&-&- & 175M & 10.3* \\
CMPP~\cite{CMPP} &92.6&73.8&-&-&82.3&57.5&-&-&- & - &1.3*  \\
JMMAC~\cite{JMMAC} &90.2&73.2&-&-&79.0&57.3&-&-&- & - & 4.0* \\
HMFT~\cite{hmft} &91.2&74.9&78.6&53.5&78.8&56.8&-&-&- & 127M & 30.2* \\
ProTrack~\cite{protrack} &82.2&69.0&-&-&79.5&59.9&53.8&49.8&42.0 & - &- \\
APFNet~\cite{APFNet} &90.5&73.7&-&-&82.7&57.9&50.0&43.9&36.2 & \textcolor{red}{15M} & 1.9* \\
CMD~\cite{cmd} &89.2&73.4&-&-&82.4&58.4&59.0&54.6&46.4  & \textcolor{green}{20M} & 30.0*\\
CAT++~\cite{cat++} &91.5&73.3&82.2&56.1&84.0&59.2&50.9&44.4&35.6  & - &14.0* \\
QueryTrack~\cite{querytrack} &92.3&75.9&-&-&84.1&60.0&66.0&-&52.0  & - & 27.0* \\
\midrule
ViPT~\cite{vipt} &90.5&75.9&-&-&83.5&61.7&65.1&61.7&52.5  & 93M & \textcolor{green}{52.9}\\
TBSI~\cite{tbsi} &89.8&75.0&\textbf{\textcolor{blue}{85.3}}&\textbf{\textcolor{blue}{62.5}}&87.1&63.7&69.2&65.7&55.6  & 202M  & 47.9\\
BAT~\cite{bat} &89.1&75.6&-&-&86.8&64.1&70.2&66.4&56.3  & 92M & \textcolor{red}{62.9} \\
GMMT~\cite{gmmt} &\textbf{\textcolor{blue}{93.6}}&\textbf{\textcolor{green}{78.5}}&-&-&87.9&64.7&\textbf{\textcolor{blue}{70.7}}&67.0&\textbf{\textcolor{blue}{56.6}}  & - & - \\
TATrack~\cite{tatrack} &-&-&\textbf{\textcolor{blue}{85.3}}&61.8&87.2&64.4&70.2&66.7&56.1  & - & 26.1*\\
UnTrack~\cite{UnTrack} &-&-&-&-&83.7&61.7&66.7&62.9&53.6  & 98M & - \\
SDSTrack~\cite{sdstrack} &87.4&75.2&-&-&84.8&62.5&66.5&62.7&53.1  & 102M & 21.9 \\
OneTracker~\cite{onetracker} &-&-&-&-&85.7&64.2&67.2&-&53.8  & 99M & - \\
STMT~\cite{STMT} &-&-&83.0&59.5&86.5&63.8&67.4&63.4&53.7  &  & 39.1* \\
TransAM~\cite{TransAM}&92.9&\textbf{\textcolor{blue}{77.4}}&-&-&87.7&\textbf{\textcolor{blue}{65.5}}&70.2&66.0&55.9  & - & - \\
$\text{SeqTrackV2-B}_\text{256}$~\cite{seqtrackv2} &-&-&-&-&\textbf{\textcolor{blue}{88.0}}&64.7&70.4&\textbf{\textcolor{blue}{67.2}}&55.8 & 92M & 30.3 \\
% $\text{SeqTrackV2-B}_\text{384}$~\cite{seqtrackv2} &-&-&-&-&90.0&66.3&71.5&67.5&56.2 \\
\midrule
\textbf{$\text{MambaVT-S}_\text{256}$}&\textbf{\textcolor{green}{94.1}}&75.3&\textbf{\textcolor{green}{88.0}}&\textbf{\textcolor{green}{63.7}}&\textbf{\textcolor{green}{88.9}}&\textbf{\textcolor{green}{65.8}}&\textbf{\textcolor{red}{73.0}}&\textbf{\textcolor{red}{69.5}}&\textbf{\textcolor{red}{57.9}} & \textcolor{blue}{29M} & \textcolor{blue}{49.7}\\
\textbf{$\text{MambaVT-M}_\text{256}$}&\textbf{\textcolor{red}{95.2}}&\textbf{\textcolor{red}{78.6}}&\textbf{\textcolor{red}{88.5}}&\textbf{\textcolor{red}{64.4}}&\textbf{\textcolor{red}{90.7}}&\textbf{\textcolor{red}{67.5}}&\textbf{\textcolor{green}{72.7}}&\textbf{\textcolor{green}{69.0}}&\textbf{\textcolor{green}{57.5}} & 79M & 31.1 \\
\bottomrule
\end{NiceTabular}
}
}
\caption{Overall performance on four prevalent RGB-T evaluation benchmarks. \textcolor{red}{Red}/\textcolor{green}{Green}/\textcolor{blue}{Blue} indicates the best/runner-up/third best results. Results are reported in percentage (\%). The number in the subscript denotes the search region resolution. ``*'' denotes that the speeds for these methods are cited from the original paper as the code and models are unavailable.}
\label{tab:comparison}
\end{table*}
%%%%%%%%%%%%%%%%%%%%%%%%%%%%%%%%%%%%%%%%%%%
\subsection{Ablation Studies}
To ensure a stable assessment of component effectiveness, unless otherwise specified, our ablation studies select results of the first stage of the MambaVT-Middle on the LasHeR dataset.

\textbf{Range of appearance and motion modeling.} The number of templates in the set and the number of frames used for trajectory inference directly impact the quality of context modeling. Therefore, we ablate how these factors affect tracking performance. As shown in \cref{tab:temnum}, performance increases steadily as the number of templates $\mathcal{M}$ increases from 1 to 7. However, larger numbers of templates lead to performance degradation, suggesting that excessively long templates may introduce noise or background interference, making it crucial to select an appropriate number of templates. Similarly, the 7-frame trajectory prompts $\mathcal{T}$ lead to the best performance. We analyze that short motion prompts are not enough to provide sufficient trajectory clues while overly long motion prompts will result in coordinate truncation and are inaccurate during fast movements.

\begin{figure}[!t]
\centering
\includegraphics[width=0.97\linewidth]{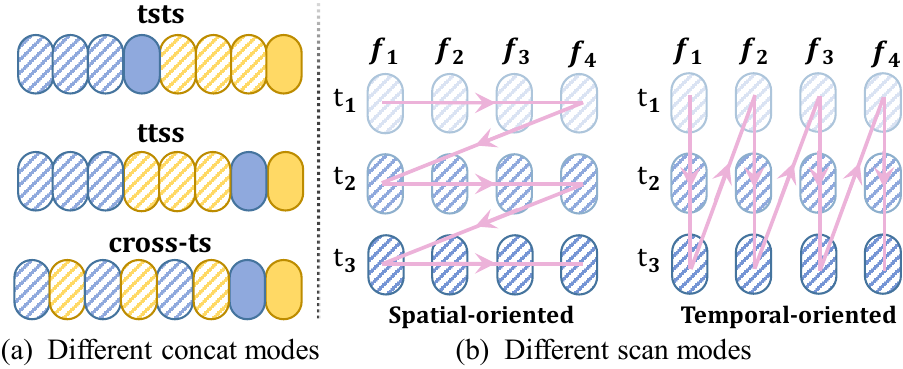}
	\caption{Various data input modes. (a) Concatenation variants of templates and search region vectors. (b) Different scan orientations of scan operator in SSM. ``t'', ``s'' and ``f'' refer to template, search region and frame, respectively.}
	\label{fig:modes}
\end{figure}

\textbf{Various data input modes.}\label{sec:modes} As shown in \cref{tab:mode}, we ablate the impact of the concatenation mode, scan mode in SSM, and frame sampling mode on the tracking performance. \cref{fig:modes} (a) illustrates the detailed concatenation modes of embedding vectors. $\textbf{\#}1$ presents better results of ``tsts'' concatenation than ``ttss'' and ``cross-ts'', indicating a large distance between the template vectors and the search area vectors imposes a significant learning burden on the model. \cref{fig:modes} (b) demonstrates scanning modes dominated by spatial orientation and temporal orientation, with results shown in lines $\textbf{\#}4$ and $\textbf{\#}5$. The spatial-oriented outcome shows superiority over the temporal-oriented one on SR by a large margin. This may be because the vector sequences generated by spatial scanning more closely align with the data distribution seen by the pre-trained model. Given a maximum sample interval, we also conduct two sample modes of multiple templates: random sample ($\textbf{\#}6$) and uniform sample ($\textbf{\#}7$). The experimental results demonstrate that the random sample is superior to the uniform sample method, which may mitigate the inductive bias.
%%%%%%%%%%%%%%%%%%%%%%%%%%%%%%%%%%%%%%%%%
\def\arraystretch{1.0}
\renewcommand{\tabcolsep}{4 pt}
\begin{table}[!t]
  \centering
  \resizebox{\linewidth}{!}{
  \begin{NiceTabular}{c|ccc||c|ccc}
    \toprule
    \#Template($\mathcal{M}$) & PR$\uparrow$ &NPR$\uparrow$ &SR$\uparrow$ & \#Coordinates($\mathcal{T}$) & PR$\uparrow$ &NPR$\uparrow$ &SR$\uparrow$  \\
    \midrule
    1 & 67.4 & 63.4 & 53.6 & 3 & 71.7 & 68.1 & 56.7 \\
    3 & 70.1 & 66.1 & 55.5 & 7 & \default{\textbf{72.7}} & \default{\textbf{69.0}} & \default{\textbf{57.5}} \\
    5 & 71.2 & 67.2 & 56.1 & 15 & 72.1 & 68.6 & 57.1  \\
    7 & \default{\textbf{71.7}} & \default{\textbf{67.6}} & \default{\textbf{56.6}} & 31 & 72.3 & 68.4 & 57.0 \\   
    9 & 71.2 & 67.2 & 56.2 & 63 & 71.4 & 68.0 & 56.5 \\
    \bottomrule
  \end{NiceTabular}
  }
  \caption{Ablation studies on the number of template images and trajectory prompts during training.}
  \label{tab:temnum}
  \vspace{-4pt}
\end{table}

\textbf{Effectiveness verification of trajectory modeling.} Incorporating motion information in the second stage required more training epochs. To fairly compare and validate the efficacy of this integration, we continue training the first stage for the same number of epochs. As shown in $\textbf{\#}8$ and $\textbf{\#}9$ of \cref{tab:mode}, merely increasing the number of training rounds does not continuously improve performance and may even lead to the drawbacks of overfitting. Conversely, the incorporation of short-term historical trajectory modeling complements long-range appearance modeling, further enhancing tracker's performance.

%%%%%%%%%%%%%%%%%%%%%%%%%%
%%%%%%%%%%%%%%%%%%%%%%%%%%
\def\arraystretch{1.0}
\renewcommand{\tabcolsep}{4 pt}
\begin{table}[ht]\footnotesize
  \centering
  \begin{NiceTabular}{c|c|ccc}
    \toprule

    \textbf{\#} & Concat  Mode & PR$\uparrow$ &NPR$\uparrow$ &SR$\uparrow$    \\
    \midrule
    1 & tsts & \default{\textbf{71.7}} & \default{\textbf{67.6}} & \default{\textbf{56.6}}  \\
    2 & ttss & 68.0 & 63.8 & 53.7 \\
    3 & cross-ts & 64.1 & 60.3 & 51.1 \\
    \midrule
    \textbf{\#} & Scan Mode & PR$\uparrow$ &NPR$\uparrow$ &SR$\uparrow$    \\ 
    \midrule
    4 & spatial-oriented & \default{\textbf{71.7}} & \default{\textbf{67.6}} & \default{\textbf{56.6}}  \\ 
    5 & temporal-oriented & 63.7 & 59.3 & 50.5 \\  
    \midrule
    \textbf{\#} & Sample Mode & PR$\uparrow$ &NPR$\uparrow$ &SR$\uparrow$    \\ 
    \midrule
    6 & random & \default{\textbf{71.7}} & \default{\textbf{67.6}} & \default{\textbf{56.6}} \\ 
    7 & uniform & 71.2 & 67.1 & 55.7 \\  
    \midrule
    \textbf{\#} & Model Variants & PR$\uparrow$ &NPR$\uparrow$ &SR$\uparrow$    \\ 
    \midrule
    8 & $+$ motion prompts & \default{\textbf{72.7}} & \default{\textbf{69.0}} & \default{\textbf{57.5}} \\ 
    9 & $+$ extra training epochs & 71.2 & 67.4 & 56.2 \\ 
    \bottomrule
  \end{NiceTabular}
  %}
  \caption{Ablation studies on various data input modes.}
  \label{tab:mode}
  \vspace{-4pt}
\end{table}

\begin{figure}[!t]
\centering
\includegraphics[width=0.485\textwidth]{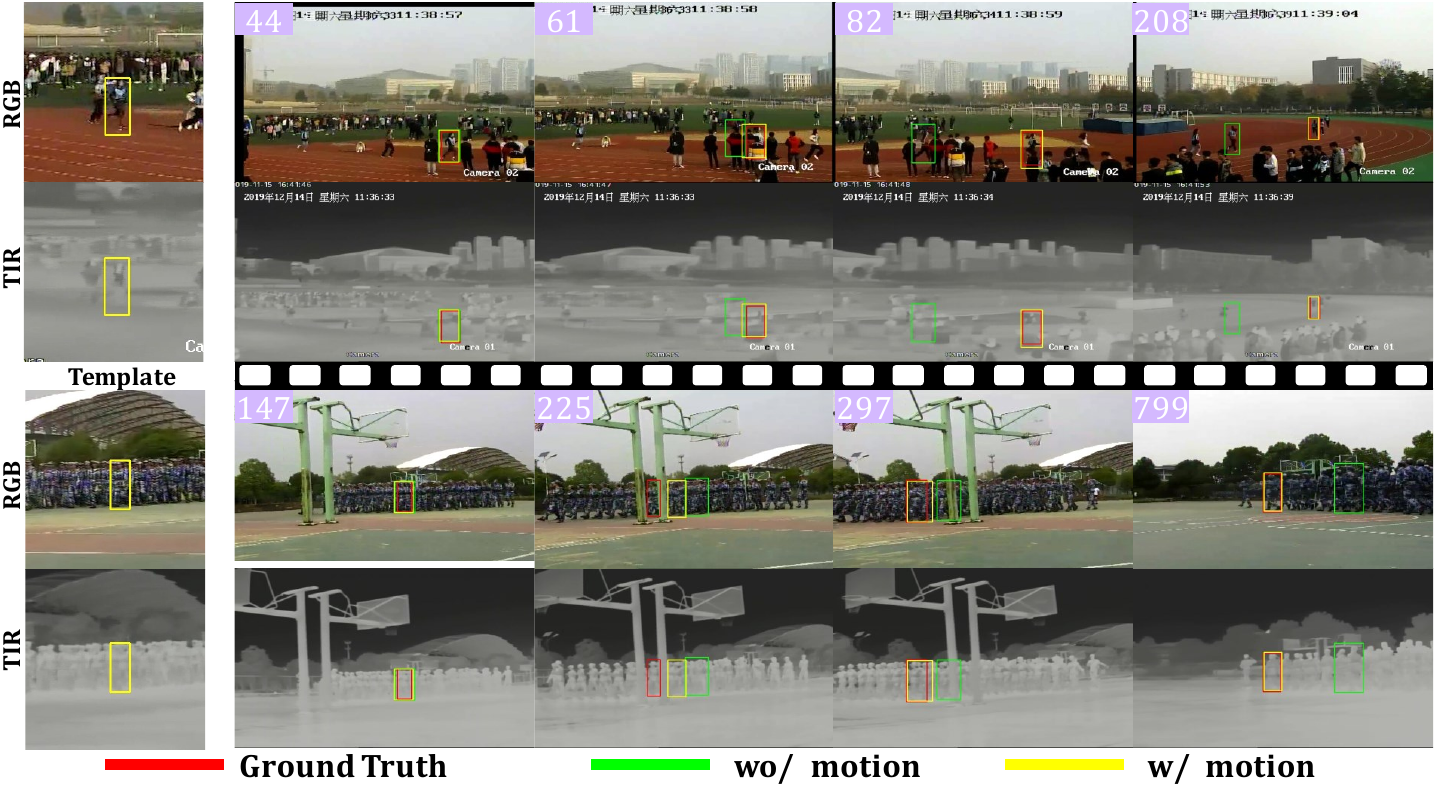}
	\caption{Qualitative comparison: before vs. after incorporating trajectory motion information}
	\label{fig:bbox}
\end{figure}

\textbf{Visualization.} To more intuitively show how the historical trajectory prompts enhance the tracker's robustness, we select the two challenging sequences and visualize the tracking results before and after incorporating motion information as illustrated in \cref{fig:bbox}. In the first video, a person is running against a clustered background. It is obvious that relying solely on appearance-based matching, the tracker drifts after passing through similar interference; however, with the aid of motion information, it can accurately follow the moving person. In the second video, when the moving target is obscured by a basketball hoop, the target's position in subsequent frames can be inferred using historical trajectories. These observations demonstrate the complementary role and effectiveness of the motion prompts in complex scenarios.

\textbf{Different Pre-training Methods.} With the common adoption of large-scale pre-trained weights for transformer-based tracking methods, we also investigate the effect of four different pre-training strategies on the tracking performance of vision Mmaba: no pre-training; ImageNet-1k\cite{deng2009imagenet} pre-trained model provided by VideoMamba\cite{li2024videomamba}; ImageNet-1k pre-trained model provided by Vim\cite{vim} and further pre-trained on the widespread RGB tracking data; ImageNet-1k pre-trained model provided by VideoMamba and further pre-trained on the RGB tracking data. As the results in \cref{tab:pretrain} show, pre-training, whether using ImageNet or RGB tracking data, is essential for guaranteed performance. Moreover, we also observe that the VideoMamba (which has the same architecture as Vim) pre-training brings better tracking performance than the Vision Mamba (Vim) pre-training, indicating its greater feature representation. We believe that with the ongoing development of the scale and paradigm of vision Mamba pre-training, akin to the ViT, our approach will gain further benefits and improvement. Note that we're using the small version since other versions of Vim are not available.
%%%%%%%%%%%%%%%%%%%%%%%%%%
%%%%%%%%%%%%%%%%%%%%%%%%%%

\def\arraystretch{1.0}
\renewcommand{\tabcolsep}{3 pt}
\begin{table}[!t]\footnotesize
  \centering
  \begin{NiceTabular}{c|c|ccc}
    \toprule
    ImageNet & RGB Tracking Data & PR$\uparrow$ &NPR$\uparrow$ &SR$\uparrow$    \\
    \midrule
    \no & \no & 45.7 & 42.4 & 36.1 \\
    VideoMamba & \no & 59.9 & 56.0 & 47.1  \\ 
    Vim & \yes & 68.4 & 64.7 & 53.9  \\ 
    VideoMamba & \yes & \default{\textbf{71.5}} & \default{\textbf{67.5}} & \default{\textbf{56.4}}  \\ 
    \bottomrule
  \end{NiceTabular}
  %}
  \caption{Ablation studies on different pre-training methods.}
  \label{tab:pretrain}
  \vspace{-4pt}
\end{table}

\textbf{Limitations and Prospects.} Currently, the parallelization of Mamba is not as advanced as that of Transformer, and the hardware lacks specific acceleration designs for Mamba, which thus limits the speed of our algorithm. However, we anticipate that continued optimizations of the Mamba architecture and hardware-specific improvement will further enhance the speed advantage of our tracker. To this end, we envisage the potential application of the newly developed Mamba2\cite{dao2024mamba2} to our framework and the experimentation with various multi-modal tracking tasks.

%%%%%%%%%%%%%%%%%%%%%%%%%%%%%%%%%%%%%%%%
%%%%%%%%%%% ablation study %%%%%%%%%%%%%
%%%%%%%%%%%%%%%%%%%%%%%%%%%%%%%%%%%%%%%%

%%%%%%%%%%% ablation study %%%%%%%%%%%%%

%%%%%%%%% CONCLUSION
\section{Conclusion}\label{sec:conclusion}
In this work, we propose Mamba-VT, a new Mamba-based RGB-T tracking framework that leverages the Mamba's exceptional property of linearly scaling with sequence length.
To comprehensively exploit mamba's contextual modeling capability with long sequence inputs, we propose the long-range cross-frame integration and short-term historical trajectory prompts in a unified and compact architecture, enabling robust RGB-T multi-modal tracking from the perspectives of global appearance modeling and local trajectory modeling.
Extensive experiments show that our MambaVT achieves state-of-the-art results on four RGB-T tracking benchmarks against with advantages of computational complexity and parameters.

\clearpage % comment it when final version
\bibliography{aaai25}

\end{document}